\documentclass[runningheads]{llncs}
\usepackage[T1]{fontenc}
\usepackage{microtype}

\def\p{{\mathrm{p}}}
\def\q{{\mathrm{q}}}

\def\bx{{\mathbf x}}

\def\bz{{\mathbf z}}

\def\bI{{\mathbf I}}

\usepackage{amsmath}
\usepackage{amsfonts}%
\usepackage{graphicx,verbatim}
\usepackage{subcaption}
\usepackage{color}

\begin{document}
\title{Bayesian generative models can flag performance loss, bias, and out-of-distribution image content}
\titlerunning{Bayesian generative models can flag issues (abbreviated)}
%

\author{Miguel López-Pérez\inst{1}\and
Marco Miani\inst{2} \and
Valery Naranjo \inst{1} \and
Søren Hauberg\inst{2} \and
Aasa Feragen\inst{2}}

\authorrunning{M. López-Pérez et al.}

\institute{Instituto Universitario de Investigación en Tecnología Centrada en el Ser Humano,
Universitat Politècnica de València, Spain \\  \email{\{mlopper3,vnaranjo\}@upv.es} \and Technical University of Denmark, Kongens Lyngby, Denmark\\  \email{\{mmia,sohau,afhar\}@dtu.dk}}


\maketitle              
\begin{abstract}
Generative models are popular for medical imaging tasks such as anomaly detection, feature extraction, data visualization, or image generation. Since they  are parameterized by deep learning models, they are often sensitive to distribution shifts and unreliable when applied to out-of-distribution data, creating a risk of, e.g. underrepresentation bias. This behavior can be flagged using uncertainty quantification methods for generative models, but their availability remains limited. We propose SLUG: A new UQ method for VAEs that combines recent advances in Laplace approximations with stochastic trace estimators to scale gracefully with image dimensionality.
We show that our UQ score -- unlike the VAE's encoder variances -- correlates strongly with reconstruction error and racial underrepresentation bias for dermatological images. We also show how pixel-wise uncertainty can detect out-of-distribution image content such as ink, rulers, and patches, which is known to induce learning shortcuts in predictive models. 


\end{abstract}

\section{Introduction}

Generative modeling is widely used in medical imaging due to its broad applications and impressive results \cite{celard2023survey}. \emph{Variational Autoencoders (VAEs)} \cite{kingmaauto,rezende14} remain popular due to their semantic, low-dimensional latent spaces, which are often used to analyze and manipulate key characteristics of high-dimensional data, e.g.\@ for data visualization \cite{ehrhardt2022autoencoders}, data generation \cite{koetzier2024generating}, and anomaly detection \cite{wijanarko2024tri}. Furthermore, VAEs can be combined with other generative models to map high-dimensional data to a lower dimensional space as in Stable Diffusion \cite{rombach2022high}.

\begin{figure}[t]
\centering
\includegraphics[width=0.6\linewidth]{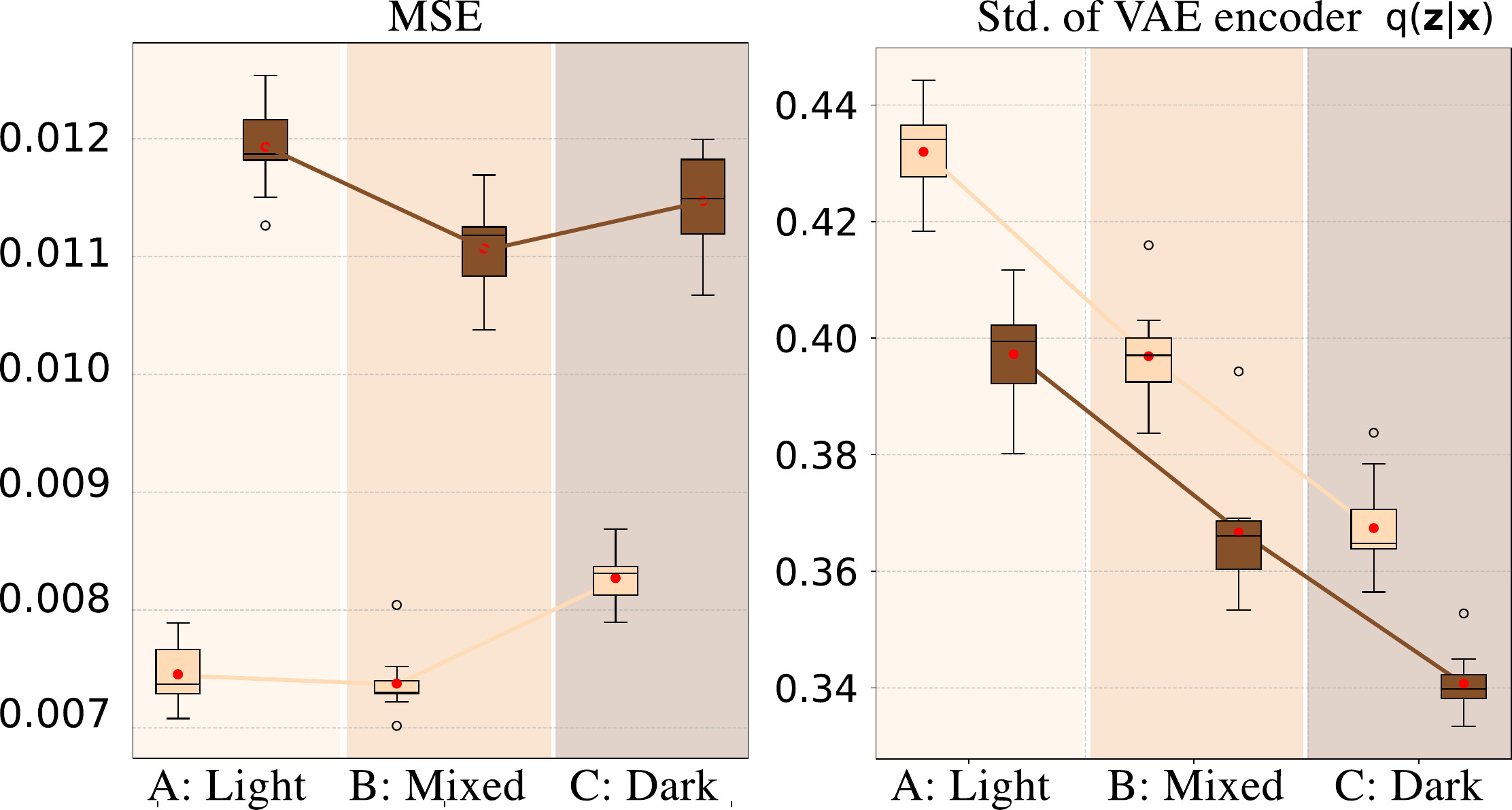}
\caption{Performance correlates with skin tone representation when training dermatological VAEs (left). However, the VAE's standard deviations do not detect this bias, illustrating why we need better UQ for generative models (right).}
\label{fig:plug}
\end{figure}

Despite these advantages, generative models are parameterized with modern Deep Neural Networks (DNNs), which struggle out-of-distribution (OOD)~\cite{nguyen2015deep}. To tackle OOD  performance in predictive models, uncertainty quantification (UQ) has emerged as an important tool~\cite{kompa2021second}, where the prediction is endowed with an associated uncertainty. This has proven useful for detecting silent failures and ensuring that unexpected outcomes do not occur. While UQ techniques have been proposed and studied with promising success for discriminative models~\cite{abdar2021review}, their applicability to generative models such as VAEs is underexplored. Fig.\ref{fig:plug} illustrates the inability of VAEs to diagnose OOD with their inherent sample-wise standard deviations, as previously reported in \cite{lopez2025generative}. The adoption of a Bayesian approach can help address this issue; however, current Bayesian generative models tend to be computationally expensive, difficult to tune, or rely on uncorrelated posterior approximations~\cite{miani2022laplacian,daxberger2019bayesian}. 

\textbf{We propose} a novel epistemic UQ method for VAEs building on the \emph{Sketched Lanczos Uncertainty (SLU)} algorithm recently proposed for discriminative models~\cite{mianisketched}. SLU computes a rank-$k$ approximation of the \emph{generalized Gauss-Newton (GNN)} matrix, which captures the epistemic uncertainty according to a Laplace approximation \cite{immer2021improving}. However, SLU scales quadratically with the output dimension, which is intractable in image-generative models. Our proposed \emph{Sketched Lanczos Uncertainty Global (SLUG)} measure overcomes this challenge using scalable stochastic trace estimators \cite{hutchinson1989stochastic} to produce a per-image score. 

\textbf{We demonstrate SLUG's ability} to detect underrepresentation bias and OOD  content in dermatological images. It is well known that dark skin tones are severely underrepresented in public datasets and that AI systems tend to reproduce and amplify this bias \cite{groh2021evaluating,lopez2025generative}. Furthermore, OOD  image content such as rulers, stickers or ink are well-known sources of shortcut learning problems as they correlate with the clinical condition~\cite{cassidy2022analysis}. Our experiments show that SLUG strongly correlates with the performance of the VAE and can serve to flag both bias and out-of-distribution image content. 

In short, \textbf{we contribute} a novel UQ method for VAEs that capture out-of-distribution data both on pixel- and image level and demonstrate its utility in flagging errors, underrepresentation bias, and out-of-distribution content using three publicly available real-world dermatology datasets.

\section{Background and related work}
\subsection{Dermatological datasets help understanding AI errors}
Racial bias is a known challenge for dermatological AI, often attributed to an underrepresentation of darker skin tones in clinical data~\cite{benmalek2024impact,kalb2023revisiting,10254437}. Previous work~\cite{lopez2025generative} indicates, as also shown in Fig.~\ref{fig:plug}, that this is not the only cause of bias: Even models trained on dark skin perform better on light skin. This highlights the need to flag high risk of underperformance, regardless of training data.

Efforts to address racial bias in dermatology have led to the creation of large-scale image datasets that include metadata on skin tone, typically measured using the \emph{Fitzpatrick Skin Type (FST)} scale, which ranges from 1 (lightest) to 6 (darkest). The \emph{Fitzpatrick17k} dataset is a key reference for assessing racial bias in dermatology, covering FST 1–6 \cite{groh2021evaluating}. However, this dataset is imbalanced, with underrepresentation of darker skin tones (FST 4–6), and studies have shown that AI models in dermatology tend to underperform on these darker skin tones \cite{groh2021evaluating}. The \emph{Diverse Dermatology Images} dataset also highlighted similar biases \cite{daneshjou2022disparities}. The \emph{PASSION} dataset  \cite{gottfrois2024passion}, focusing on individuals from Sub-Saharan countries, includes FST 3–6 and aims to address the limitations of datasets that primarily focus on lighter skin tones.

Another popular dermatological dataset is the \emph{ISIC} challenge dataset \cite{codella2018skin,tschandl2018ham10000,combalia2019bcn20000} with skin lesions and cancer diagnoses. As ISIC does not report skin color and primarily contains dermoscopic images of lighter skin tones, it is not useful for assessing bias, but it has been used to highlight another type of AI error: Special image content, such as rulers, ink, and patches, directly correlates with the diagnostic label~\cite{nauta2021uncovering}. These non-lesion features can be used as shortcuts by diagnostic models, and it would be useful for an OOD  method to highlight them \cite{cassidy2022analysis}.\looseness=-1

\subsection{Uncertainty quantification in VAEs}
VAEs can estimate the density of high-dimensional data, such as images. However, the predicted uncertainty depends on the quality of the likelihood estimated by the DNNs parameterizing the VAE, which are poorly calibrated and often inaccurate. To address this, some UQ methods have been proposed. Deep ensembles are considered the gold standard but are computationally expensive \cite{lakshminarayanan2017simple}, while Monte Carlo dropout is easy to implement but has poor empirical performance \cite{kanwal2024you}.

Some studies have made efforts toward a scalable full Bayesian approach, where a distribution over the weights of the VAE is computed. Daxberger et al.\@ \cite{daxberger2019bayesian} proposed a Bayesian VAE that replaced the decoder point estimate with samples from the posterior using stochastic gradient Markov Chain Monte Carlo (MCMC). Miani et al.\@  \cite{miani2022laplacian} introduced a Laplace approximation to construct a variational distribution over the weights of a VAE. These two works demonstrated an improved capability to detect OOD data with VAEs. Our work aligns more closely with the latter, as we aim to estimate post-hoc an approximation of the Hessian of the loss function to compute uncertainty.

\section{Method}
Let $f_{\phi,\theta}:\mathbb{R}^{W\times H \times C}\rightarrow \mathbb{R}^{W\times H \times C}$ be a VAE decoder trained to generate images of size $W \times H \times C$, parameterized by ${\phi,\theta}\in\mathbb{R}^p$. We denote by $\mathbf{J}_{\{\phi,\theta\}}(x)=\nabla_{\phi,\theta} f_{\phi,\theta}(x)\in\mathbb{R}^{({W\times H \times C)\times p}}$ its Jacobian w.r.t.\@ the parameters, evaluated at a given image $x\in \mathbb{R}^{W\times H \times C}$. The \emph{Generalized Gaussian-Newton (GNN)} matrix is then defined as
\begin{equation}\label{GGN}
    \mathbf{G_{\{\phi,\theta\}}}=\sum^n_{i=1}\mathbf{J}_{\{\phi,\theta\}}(x_i)^{T}\mathbf{H}(x_i)\mathbf{J}_{\{\phi,\theta\}}(x_i),
\end{equation}
where $\mathbf{H}(x_i)$ is the Hessian of the loss with respect to the neural network output. 

The GGN commonly appears as the inverse covariance of the \emph{linearized Laplace approximation (LLA)} to the true posterior \cite{immer2021improving}. Currently, LLA is the most promising Bayesian posterior approximation \cite{immer2021improving}, but it is, unfortunately, intractable for generative models as its computational cost scales quadratically with the generated data dimension, specifically \@ $\mathcal{O}((WHC)^2 p)$ for a network with $p$ parameters.

Recently, Miani et al.\@ \cite{mianisketched} developed a sketching-based algorithm to evaluate the associated predictive uncertainty, which scales logarithmically with $p$. The resulting \emph{Sketched Lanczos Uncertainty (SLU)} algorithm, however, still scales quadratically with the image dimension, making it impractical for VAEs. Our approach extends SLU to scale gracefully to large images.

\subsection{Generative model: Variational Autoencoder}

A VAE is a probabilistic generative model with an autoencoder-like architecture combining a \textit{stochastic encoder} $\q_{\phi}(\bx|\bz)$ (parameterized by $\phi$), and a \textit{stochastic decoder} $\p_\theta(\bx|\bz)$ (parameterized by $\theta$). 
VAEs utilize amortized variational inference to learn these probability distributions by parameterizing them with DNNs, thus, the uncertainty of this probability distribution does not generalize well out of distribution. 
 
The optimal variational parameters of the probability distributions, $\{\theta,\phi\}$, are computed by maximizing the marginal log-likelihood lower bound (ELBO):
\begin{equation} \label{eq:elbo}
    \textrm{ELBO}(\theta,\phi;\mathbf{x}) = \mathbb{E}_{q_\phi(\mathbf{z}|\mathbf{x})}\left[\log p_\theta(\mathbf{x}|\mathbf{z}) \right] - \textrm{KL}(q_\phi(\mathbf{z}|\bx)||p(\mathbf{z})),
\end{equation}
where  $\textrm{KL}(q_\phi(\mathbf{z}|\bx)||p(\mathbf{z}))$ is the Kullback-Leibler (KL) divergence between the approximate posterior $q_\phi(\mathbf{z}|\bx)$ and the prior $\mathcal{N}(z|0,\bI)$. The first term in Eq.~\ref{eq:elbo} corresponds to the mean square error (MSE) because we utilize a Gaussian likelihood. To estimate the distribution $\p_\theta(\bz|\bx)$, we utilize the reparameterization trick to obtain samples of the latent variable $\bz=\mu(\bx) + \epsilon \odot \sigma(\bx)$, $\epsilon\sim\mathcal{N}(0,\bI)$.

To ensure high-fidelity images, we train the VAE with a perceptual loss \cite{hou2017deep}. Instead of summing per-pixel losses, the perceptual loss is computed in the feature space of a pretrained neural network $\Phi$. The idea is that the feature space loss is perceptually more meaningful than the per-pixel MSE. The total loss of the VAE is then
 \begin{equation}\label{eq:perploss}
     \mathcal{L(\{\phi,\theta\})} = \mathrm{ELBO} + \frac{1}{2C^lW^lH^l}\sum^{C^l}_{c=1}\sum^{W^l}_{w=1}\sum^{H^l}_{h=1}(\Phi(\bx)^l_{c,w,h} - \Phi(f_\theta(x))_{c,w,h})^2,
 \end{equation}
 where $C^l, H^l, W^l$ are the channels, height, and width of the $l$-th feature map of the network, respectively.

\subsection{Scaling to VAEs: Sketched Lanczos Uncertainty Global score (SLUG)}
We based our proposed score on the SLU algorithm \cite{miani2022laplacian}, which we briefly review. Let $\mathbf{U}$ denote the matrix containing the leading eigenvectors of the GGN \eqref{GGN}, then the SLU approximates the predictive variance of the linearized Laplace approximation with $\bI - \mathbf{U} \mathbf{U}^{\top}$ is covariance, i.e.
\begin{align}
  \textrm{SLU}(x)
    &=\text{Tr}(\mathbf{J}_{\theta^*}(x)\ (\bI - \mathbf{UU}^{\top})\ \mathbf{J}_{\theta^*}(x)^{\top}).
\end{align}
SLU approximates this predictive uncertainty using several tricks from randomized numerical linear algebra. These are essential to build a scalable approximation, but, from our perspective, they can be treated as `black box' and we refer the reader to the original paper for the details \cite{mianisketched}.

Unfortunately, even SLU does not scale to neural networks with high-dimensional outputs like those in generative models. Producing one predictive variance per generated pixel requires $\mathcal{O}(WHC)$ SLU invocations, which is practically prohibitive.

Our main interest is in measuring a scalar uncertainty score for a generated image, and we choose the sum of per-pixel predictive variances, which we denote the \emph{Sketched Lanczos Uncertainty Global (SLUG)} score,
\begin{align}
    \textrm{SLUG}(x) = \sum_{w,h,c}\textrm{SLU}(x_{w,h,c}).
\end{align}
As a naive implementation requires $\mathcal{O}(WHC)$ SLU invocations, we propose to use a stochastic trace estimator \cite{hutchinson1989stochastic} to approximate the sum,
\begin{align}
   \textrm{SLUG}(x)
    &=\text{Tr}(\mathbf{J}_{\theta^*}(x)\cdot(\bI - \mathbf{UU}^{\top})\cdot\mathbf{J}_{\theta^*}(x)^{\top}) \\
    &=\mathbb{E}_{\epsilon\sim\p(\epsilon)}
    \left[
    \epsilon \mathbf{J}_{\theta^*}(x)\ (\bI - \mathbf{UU}^{\top})\ \mathbf{J}_{\theta^*}(x)^{\top}\epsilon^{\top}
    \right] \\ \nonumber
    &\approx\frac{1}{S}\sum_{s=1}^S
      \epsilon_s \mathbf{J}_{\theta^*}(x) \ (\bI - \mathbf{UU}^{\top})\ \mathbf{J}_{\theta^*}(x)^{\top}\epsilon_s^{\top},
\end{align}
where $\epsilon_s \sim \mathcal{N}(0,\bI)$. This can be implemented using only $S$ invocations of SLU. A near-identical estimator of per-pixel variances can be constructed by replacing the stochastic trace estimate with a stochastic estimate of the matrix diagonal, which, again, can be realized using $S$ SLU invocations.

\section{Experiments}
\subsubsection{Datasets.} In our experiments, we use three real-world datasets from dermatology. For all experiments, we train a VAE on the \textbf{Fitzpatrick17k dataset} \cite{groh2021evaluating}, which is a large publicly available dataset under license (CC BY-NC-SA), consisting of 16,577 images from two dermatological atlases with corresponding FST labels. To assess the influece of skin tone representation in the training set, we divide this dataset in lighter skin tones (FST 1-2) and darker skin tones (FST 5-6), and independently sample three different datasets for each run, which are defined as follows: The `Dataset A -- Light' has 100\% of lighter images, the `Dataset B -- Mixed' set has 50/50, and the `Dataset C -- Black' set has 100\% of darker images, with total size of $1668$ samples each of them. We also sample two test sets for lighter and darker skin tones with $512$ samples each. We utilize the \textbf{PASSION dataset} \cite{gottfrois2024passion}, publicly available under license  (CC BY-NC), for external validation to assess the racial bias of the model. This publicly available dataset includes 4,901 dermatology images from different Subsaharian countries with darker skin tones (from FST3 to FST6). We utilize the \textbf{ISIC dataset} \cite{codella2018skin,tschandl2018ham10000,combalia2019bcn20000}, publicly available, where one subset is under license (CC BY-NC), and the remainder under (CC0), to validate the pixelwise OOD detection in dermatology. This dataset contains dermatological lesions as well as some OOD elements. 

\subsubsection{VAE architecture and implementation details.} The encoder and decoder are composed of residual blocks, along with down-sampling and up-sampling paths, respectively. Each convolutional block employs batch normalization and ELU activation. We train the models for 1000 epochs, after which we observe convergence in the training loss. The optimizer used was Adam, with a cosine decay scheduler. Training was performed using mini-batches of 64 samples. We resize all images to $128\times128$ and perform 10 independent runs. For the uncertainty quantification score, we utilize $S=500$ samples. The experiments were implemented with JAX 4.37, Flax 0.10.2, CUDA 12.7, and executed on an NVIDIA Tesla V100 with 32 GB of memory.

\subsubsection{Racial bias in the Fitzpatrick17k dataset.}
\begin{figure}[t]
    \centering
    \includegraphics[width=\linewidth]{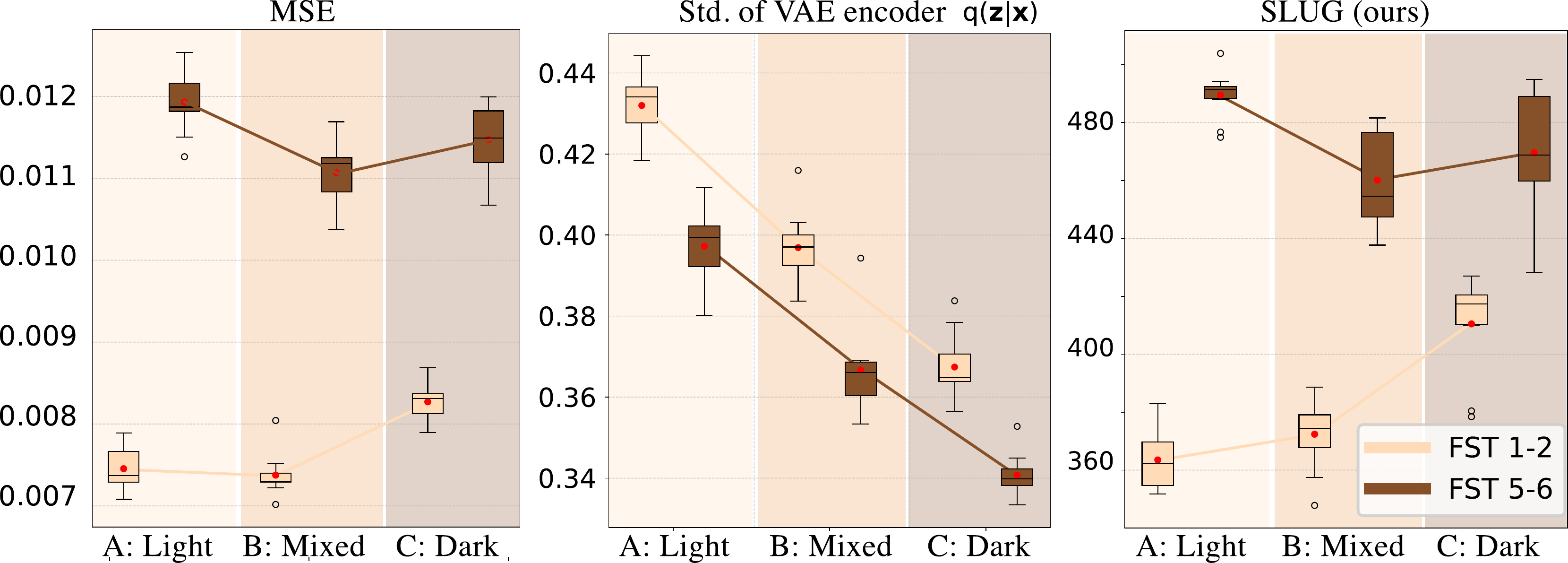}
    \caption{On Fitzpatrick17k, the performance on light and dark skin tones changes with their representation. The VAE encoder uncertainty is a poor indicator, while SLUG follows performance across groups and training scenarios.}
    \label{fig:metrics}
\end{figure}

On Dataset B -- Mixed, we compute the correlation between our proposed SLUG score and the VAE's performance error (see Fig.~\ref{fig:correlation}). These are strongly correlated, meaning that the SLUG score serves as a proxy for detecting errors. Fig.~ \ref{fig:metrics} demonstrates that this relation also remains on the skin tone subgroup level, meaning that SLUG also captures racial bias --  in strong contrast to the VAE's latent uncertainty.

\subsubsection{Racial bias in the external PASSION dataset.}
\begin{figure}[t]
        \centering
        \includegraphics[width=\linewidth]{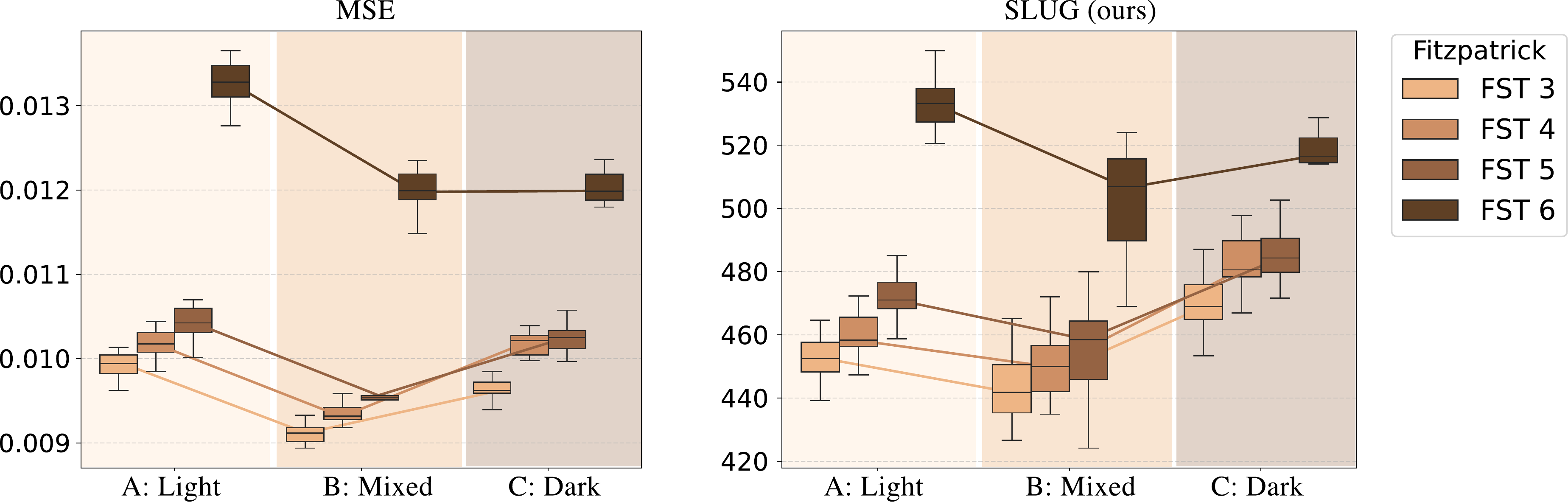}
        \caption{On the external PASSION dataset, we see again how reduced MSE is flagged by increased SLUG uncertainty across dark skin tone groups.}
        \label{fig:passion}
        \end{figure}

We calculate the MSE and SLUG score in the external dataset PASSION for each training configuration. In this dataset, we find a similar bias as the performance worsens as the skin tone is darker (see Fig. \ref{fig:passion}).  We observe that the SLUG score is also able to capture this racial bias in the external dataset.

\subsubsection{OOD image content detection in the external ISIC dataset.}
\begin{figure}
    \centering
    \includegraphics[width=\linewidth]{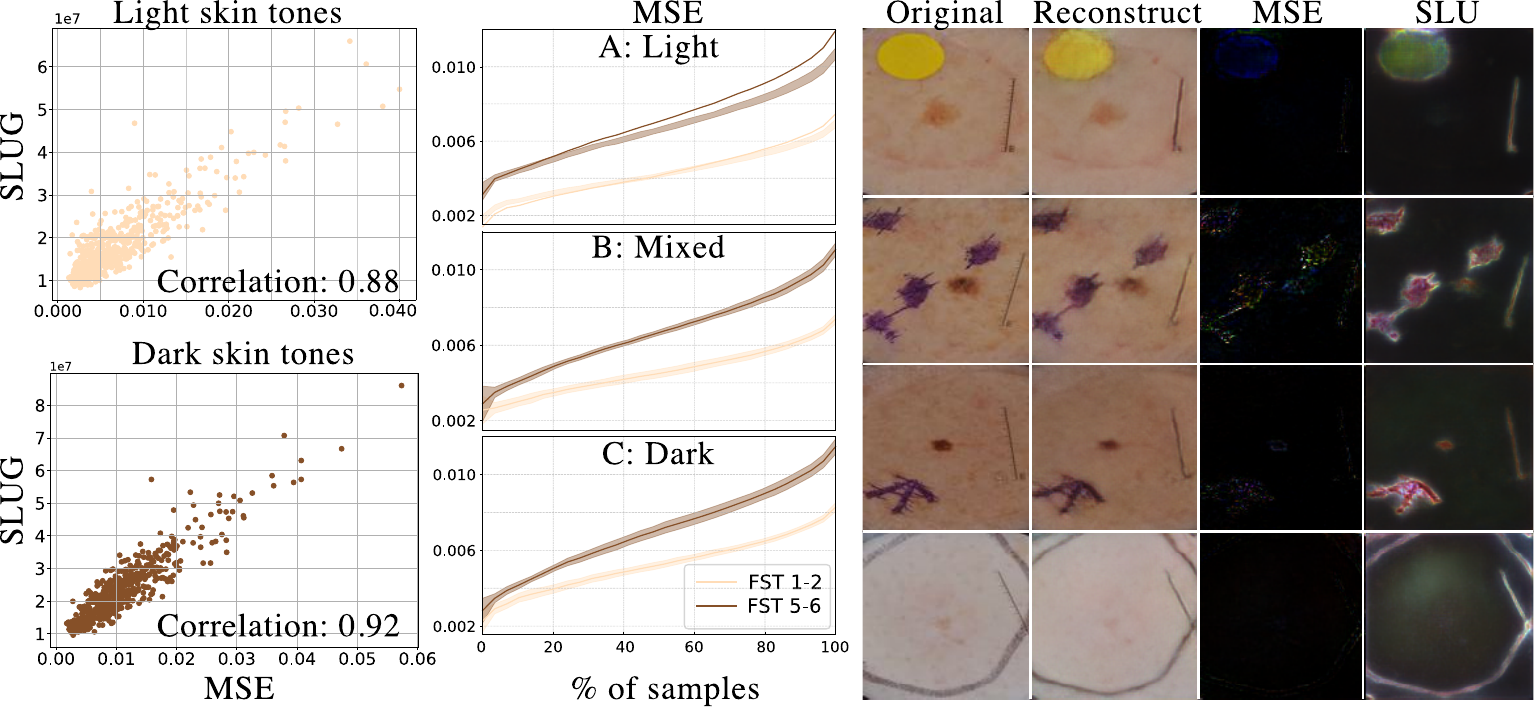}
    \caption{\emph{Left:} On Fitzpatrick17k (Dataset B -- Mixed), our SLUG score strongly correlates with the MSE. \emph{Center:} On Fitzpatrick17k, removing samples with higher uncertainty results in consistent improvements. \emph{Right:} On ISIC, the VAE reconstructs OOD  data, but SLU detects the OOD  content in dermatological images.\looseness=-1}
    \label{fig:correlation}
\end{figure}

We qualitatively analyze the ability of our SLUG score to detect OOD content in dermatology images, see four illustrative examples in Fig.~\ref{fig:correlation}. The MSE and UQ maps are computed by normalizing the pixel-wise uncertainties to the [0-1] interval and displayed as RGB images. Note that the rulers, ink and patch are all highlighted well. An interesting observation is that in B and C, the lesion also exhibits high uncertainty, but lower than that of the OOD  objects. On the other hand, as the VAE reconstructs the OOD  objects well, the MSE is unable to flag them.

\section{Discussion and Conclusion}
This work highlights the urgent need for precise and scalable UQ for generative models. Despite the widespread use of generative AI, we still lack a reliable mechanism to ensure its trustworthiness. We demonstrate that epistemic UQ can warn of performance loss, highlight biases, and detect OOD  image content.

\textbf{SLUG detects bias.} As shown by~\cite{lopez2025generative}, generative models such as VAEs can suffer from racial bias in dermatology imaging, and the intrinsic latent standard deviation of the VAE is unable to capture this bias. Our proposed SLUG score, conversely, strongly correlates with VAE performance in reconstructing images, also on skin tone subgroups in the Fitzpatrick17k and PASSION datasets. This shows the SLUG score's ability to capture racial bias. 

Our results insinuate that this bias arises partially from subgroup representation -- but representation does not explain all: Both MSE error and SLUG uncertainty are consistently higher for dark skin tones even when the VAE is trained exclusively on images of dark skin. Moreover, when training on exclusively dark skin images, performance is lower for all skin tone groups in both test sets than when training on a balanced selection of skin tones. This suggests that something else is at play. We hypothesize that a potential explanation could be higher variability and hence more complex image distribution for the dark skinned images -- this could e.g.\@ happen if inclusion criteria are relaxed for a potentially harder-to-recruit black population. Discovering the true causes of bias therefore remains an important open challenge. Meanwhile, we note that even if we do not know the complete cause of bias, our SLUG score is still able to warn against it when deployed in clinical practice.

\textbf{SLUG detects OOD  image content.} We have demonstrated that our epistemic UQ, when applied pixelwise, is useful for detecting OOD  content within dermatology images. These OOD  elements could include items such as a ruler, which does not contain diagnostic information but correlates with the diagnosis and may serve as a shortcut for discriminative models. The uncertainty map's ability to highlight this OOD  image content could prove useful also for avoiding shortcut problems. 

The main \textbf{limitation} of our work is that the computation of pixelwise uncertainty does not scale well, posing a bottleneck for high-resolution or 3D images. In summary, however, our model and results \textbf{highlight that UQ for generative models show excellent potential for safeguarding their utilization.}

\begin{credits}

\subsubsection{\ackname}
This research is partly funded by José Castillejo mobility scholarship (CAS23/00387) granted to MLP by the Spanish Government to visit DTU, MLP has also been supported by MICIU/AEI/10.13039/501100011033 and the European Union’s “NextGenerationEU”/PRTR through grant JDC2022-048318-I and by ``Ayuda a Primeros"
Proyectos de Investigacion (PAID-06-22), Vicerrectorado de Investigación de la Universitat Politécnica de València (UPV). VN was supported by the Spanish Ministry of Economy and Competitiveness
through the project PID2022-140189OB-C21 (ASSIST) and CIPROM/2022/20 (PROMETEO) by GVA.  The work was partly funded by the Novo Nordisk Foundation through the Center for Basic Machine Learning Research in Life Science (NNF20OC0062606). AF was supported by research grant (0087102) from the Novo Nordisk Foundation. SH was supported by a research grant (42062) from VILLUM FONDEN and from the European Research Council (ERC) under the European Union’s Horizon programme (grant agreement 101125993).
\subsubsection{\discintname}
The authors have no competing interests to declare that are
relevant to the content of this article.
\end{credits}


\bibliographystyle{splncs04}
\bibliography{ref}

\begin{thebibliography}{10}
\providecommand{\url}[1]{\texttt{#1}}
\providecommand{\urlprefix}{URL }
\providecommand{\doi}[1]{https://doi.org/#1}

\bibitem{abdar2021review}
Abdar, M., Pourpanah, F., Hussain, S., Rezazadegan, D., Liu, L., Ghavamzadeh, M., Fieguth, P., Cao, X., Khosravi, A., Acharya, U.R., et~al.: A review of uncertainty quantification in deep learning: Techniques, applications and challenges. Information fusion  \textbf{76},  243--297 (2021)

\bibitem{10254437}
Almuzaini, A.A., Dendukuri, S.K., Singh, V.K.: Toward fairness across skin tones in dermatological image processing. In: 2023 IEEE 6th International Conference on Multimedia Information Processing and Retrieval (MIPR). pp.~1--7 (2023)

\bibitem{benmalek2024impact}
Benmalek, A., Cintas, C., Tadesse, G.A.: Impact of skin tone diversity on out-of-distribution detection methods in dermatology. In: International Conference on Medical Image Computing and Computer-Assisted Intervention (2024)

\bibitem{cassidy2022analysis}
Cassidy, B., Kendrick, C., Brodzicki, A., Jaworek-Korjakowska, J., Yap, M.H.: Analysis of the isic image datasets: Usage, benchmarks and recommendations. Medical image analysis  \textbf{75},  102305 (2022)

\bibitem{celard2023survey}
Celard, P., Iglesias, E.L., Sorribes-Fdez, J.M., Romero, R., Vieira, A.S., Borrajo, L.: A survey on deep learning applied to medical images: from simple artificial neural networks to generative models. Neural Computing and Applications  \textbf{35}(3),  2291--2323 (2023)

\bibitem{codella2018skin}
Codella, N.C., Gutman, D., Celebi, M.E., Helba, B., Marchetti, M.A., Dusza, S.W., Kalloo, A., Liopyris, K., Mishra, N., Kittler, H., et~al.: Skin lesion analysis toward melanoma detection: A challenge at the 2017 international symposium on biomedical imaging (isbi), hosted by the international skin imaging collaboration (isic). In: 2018 IEEE 15th international symposium on biomedical imaging (ISBI 2018). pp. 168--172. IEEE (2018)

\bibitem{combalia2019bcn20000}
Combalia, M., Codella, N.C., Rotemberg, V., Helba, B., Vilaplana, V., Reiter, O., Carrera, C., Barreiro, A., Halpern, A.C., Puig, S., et~al.: Bcn20000: Dermoscopic lesions in the wild. arXiv preprint arXiv:1908.02288  (2019)

\bibitem{daneshjou2022disparities}
Daneshjou, R., Vodrahalli, K., Novoa, R.A., Jenkins, M., Liang, W., Rotemberg, V., Ko, J., Swetter, S.M., Bailey, E.E., Gevaert, O., et~al.: Disparities in dermatology ai performance on a diverse, curated clinical image set. Science advances  \textbf{8}(31),  eabq6147 (2022)

\bibitem{daxberger2019bayesian}
Daxberger, E., Hern{\'a}ndez-Lobato, J.M.: Bayesian variational autoencoders for unsupervised out-of-distribution detection. arXiv preprint arXiv:1912.05651  (2019)

\bibitem{ehrhardt2022autoencoders}
Ehrhardt, J., Wilms, M.: Autoencoders and variational autoencoders in medical image analysis. In: Biomedical Image Synthesis and Simulation, pp. 129--162. Elsevier (2022)

\bibitem{gottfrois2024passion}
Gottfrois, P., Gr{\"o}ger, F., Andriambololoniaina, F.H., Amruthalingam, L., Gonzalez-Jimenez, A., Hsu, C., Kessy, A., Lionetti, S., Mavura, D., Ng’ambi, W., et~al.: Passion for dermatology: Bridging the diversity gap with pigmented skin images from sub-saharan africa. In: International Conference on Medical Image Computing and Computer-Assisted Intervention. pp. 703--712. Springer (2024)

\bibitem{groh2021evaluating}
Groh, M., Harris, C., Soenksen, L., Lau, F., Han, R., Kim, A., Koochek, A., Badri, O.: Evaluating deep neural networks trained on clinical images in dermatology with the fitzpatrick 17k dataset. In: Proceedings of the IEEE/CVF Conference on Computer Vision and Pattern Recognition. pp. 1820--1828 (2021)

\bibitem{hou2017deep}
Hou, X., Shen, L., Sun, K., Qiu, G.: Deep feature consistent variational autoencoder. In: 2017 IEEE winter conference on applications of computer vision (WACV). pp. 1133--1141. IEEE (2017)

\bibitem{hutchinson1989stochastic}
Hutchinson, M.F.: A stochastic estimator of the trace of the influence matrix for laplacian smoothing splines. Communications in Statistics-Simulation and Computation  \textbf{18}(3),  1059--1076 (1989)

\bibitem{immer2021improving}
Immer, A., Korzepa, M., Bauer, M.: Improving predictions of bayesian neural nets via local linearization. In: International conference on artificial intelligence and statistics. pp. 703--711. PMLR (2021)

\bibitem{kalb2023revisiting}
Kalb, T., Kushibar, K., Cintas, C., Lekadir, K., Diaz, O., Osuala, R.: Revisiting skin tone fairness in dermatological lesion classification. In: Workshop on Clinical Image-Based Procedures. pp. 246--255. Springer (2023)

\bibitem{kanwal2024you}
Kanwal, N., L{\'o}pez-P{\'e}rez, M., Kiraz, U., Zuiverloon, T.C., Molina, R., Engan, K.: Are you sure it’s an artifact? artifact detection and uncertainty quantification in histological images. Computerized Medical Imaging and Graphics  \textbf{112} (2024)

\bibitem{kingmaauto}
Kingma, D.P., Welling, M.: Auto-encoding variational bayes. In: International Conference on Learning Representations (ICLR) (2014)

\bibitem{koetzier2024generating}
Koetzier, L.R., Wu, J., Mastrodicasa, D., Lutz, A., Chung, M., Koszek, W.A., Pratap, J., Chaudhari, A.S., Rajpurkar, P., Lungren, M.P., et~al.: Generating synthetic data for medical imaging. Radiology  \textbf{312}(3),  e232471 (2024)

\bibitem{kompa2021second}
Kompa, B., Snoek, J., Beam, A.L.: Second opinion needed: communicating uncertainty in medical machine learning. NPJ Digital Medicine  \textbf{4}(1), ~4 (2021)

\bibitem{lakshminarayanan2017simple}
Lakshminarayanan, B., Pritzel, A., Blundell, C.: Simple and scalable predictive uncertainty estimation using deep ensembles. Advances in neural information processing systems  \textbf{30} (2017)

\bibitem{lopez2025generative}
L{\'o}pez-P{\'e}rez, M., Hauberg, S., Feragen, A.: Are generative models fair? a study of racial bias in dermatological image generation. arXiv:2501.11752  (2025)

\bibitem{mianisketched}
Miani, M., Beretta, L., Hauberg, S.r.: Sketched lanczos uncertainty score: a low-memory summary of the fisher information. In: Globerson, A., Mackey, L., Belgrave, D., Fan, A., Paquet, U., Tomczak, J., Zhang, C. (eds.) Advances in Neural Information Processing Systems. vol.~37, pp. 23123--23154 (2024)

\bibitem{miani2022laplacian}
Miani, M., Warburg, F., Moreno-Mu{\~n}oz, P., Skafte, N., Hauberg, S.: Laplacian autoencoders for learning stochastic representations. Advances in Neural Information Processing Systems  \textbf{35},  21059--21072 (2022)

\bibitem{nauta2021uncovering}
Nauta, M., Walsh, R., Dubowski, A., Seifert, C.: Uncovering and correcting shortcut learning in machine learning models for skin cancer diagnosis. Diagnostics  \textbf{12}(1), ~40 (2021)

\bibitem{nguyen2015deep}
Nguyen, A., Yosinski, J., Clune, J.: Deep neural networks are easily fooled: High confidence predictions for unrecognizable images. In: Proceedings of the IEEE conference on computer vision and pattern recognition. pp. 427--436 (2015)

\bibitem{rezende14}
Rezende, D.J., Mohamed, S., Wierstra, D.: Stochastic backpropagation and approximate inference in deep generative models. In: Xing, E.P., Jebara, T. (eds.) Proceedings of the 31st International Conference on Machine Learning. Proceedings of Machine Learning Research, vol.~32, pp. 1278--1286. PMLR (2014)

\bibitem{rombach2022high}
Rombach, R., Blattmann, A., Lorenz, D., Esser, P., Ommer, B.: High-resolution image synthesis with latent diffusion models. In: Proceedings of the IEEE/CVF conference on computer vision and pattern recognition. pp. 10684--10695 (2022)

\bibitem{tschandl2018ham10000}
Tschandl, P., Rosendahl, C., Kittler, H.: The ham10000 dataset, a large collection of multi-source dermatoscopic images of common pigmented skin lesions. Scientific data  \textbf{5}(1), ~1--9 (2018)

\bibitem{wijanarko2024tri}
Wijanarko, H., Calista, E., Chen, L.F., Chen, Y.S.: Tri-vae: Triplet variational autoencoder for unsupervised anomaly detection in brain tumor mri. In: Proceedings of the IEEE/CVF Conference on Computer Vision and Pattern Recognition. pp. 3930--3939 (2024)

\end{thebibliography}
\end{document}